\documentclass{article}
\usepackage{spconf,amsmath,epsfig}
\usepackage[utf8]{inputenc} 
\usepackage[T1]{fontenc}    
\usepackage{hyperref}       
\usepackage{url}            
\usepackage{booktabs}       
\usepackage{amsfonts}       
\usepackage{amsmath}
\usepackage{amssymb}
\usepackage{nicefrac}       
\usepackage{microtype}      
\usepackage{xcolor}         
\usepackage{xspace}
\usepackage{multirow}
\usepackage{adjustbox}

\usepackage{graphicx}
\usepackage{colortbl} 

\usepackage{amsmath}
\usepackage{booktabs}
\usepackage{xspace}
\usepackage{amssymb}
\usepackage[utf8]{inputenc} 
\usepackage[T1]{fontenc}    
\usepackage{hyperref}       
\usepackage{url}            
\usepackage{booktabs}       
\usepackage{amsfonts}       
\usepackage{nicefrac}       
\usepackage{microtype}      
\usepackage{lipsum}
\usepackage{fancyhdr}       
\usepackage{graphicx}       
\usepackage{enumitem}
\usepackage{multirow}

\newcommand{\sims}{\texttt{Sims}\xspace}

\title{\sims: An Interactive Tool for Geospatial Matching and Clustering}

\name{\begin{tabular}{c}
Akram Zaytar\sthanks{Corresponding author: \texttt{akramzaytar@microsoft.com}}\textsuperscript{1}, Girmaw Abebe Tadesse\textsuperscript{1}, Caleb Robinson\textsuperscript{1}, Eduardo G. Bendito\textsuperscript{2}, Medha Devare\textsuperscript{2} \\
Meklit Chernet\textsuperscript{2}, Gilles Q. Hacheme\textsuperscript{1}, Rahul Dodhia\textsuperscript{1}, Juan M. Lavista Ferres\textsuperscript{1}
\end{tabular}}

\address{\textsuperscript{1}Microsoft AI for Good Research Lab  \\ \textsuperscript{2}CGIAR}

\begin{document}
\maketitle

\begin{abstract}
Acquiring, processing, and visualizing geospatial data requires significant computing resources, especially for large spatio-temporal domains. This challenge hinders the rapid discovery of predictive features, which is essential for advancing geospatial modeling. To address this, we developed Similarity Search (\sims), a no-code web tool that allows users to perform clustering and similarity search over defined regions of interest using Google Earth Engine as a backend. \sims is designed to complement existing modeling tools by focusing on feature exploration rather than model creation. We demonstrate the utility of \sims through a case study analyzing simulated maize yield data in Rwanda, where we evaluate how different combinations of soil, weather, and agronomic features affect the clustering of yield response zones. \sims is open source and available at \url{https://github.com/microsoft/Sims}.
\end{abstract}


\section{Introduction}\label{sec:introduction}

In geospatial analysis, acquiring, processing, and visualizing spatial data is time-consuming, particularly when dealing with datasets covering large spatio-temporal domains~\cite{bill2022geospatial}. This makes it difficult to explore different layers and identify the most useful ones for a given task (e.g., targeting geographic interventions). Consequently, in many workflows, data products are selected based on domain expertise with limited exploratory analysis. Similar bottlenecks arise when building data processing pipelines. As the spatio-temporal domain expands, the exploration process slows due to longer acquisition and processing times, coupled with a lack of immediate visual feedback, making it difficult to iterate quickly.

Beyond these computational challenges, geospatial data presents unique statistical concerns~\cite{rolf2024mission}. Satellite imagery and other spatial data typically violate the independent and identically distributed assumption due to spatial autocorrelation—nearby locations tend to be more similar than distant ones. This spatial dependency complicates traditional methods for sampling, model validation, and experimental design, often leading to over-confident model evaluations or biased outcomes. While clustering techniques can help by grouping similar regions and enabling cluster-level sampling, the resource-intensive nature of geospatial data makes it difficult to efficiently implement these approaches in practice. Lastly, there are capacity hurdles that exclude those who may need to harness the power of geospatial data, but they are unable to do so because they do not have the required expertise.

Several tools have been developed to address these challenges. For instance, \texttt{geemap}~\cite{geemap} and \texttt{Earth Map}~\cite{morales2023earth} provide interactive mapping and data exploration capabilities using Google Earth Engine (GEE)~\cite{gorelick2017google} as a backend. These tools have made significant strides in improving accessibility to spatial data layers. However, there remains a need for a no-code solution that focuses on the discovery of geospatial features for modeling tasks. This need is especially acute for data scientists who have limited experience with Geographic Information Systems (GIS) methodologies and software.

\begin{figure*}[h!]
    \centering
    \includegraphics[width=0.8\textwidth]{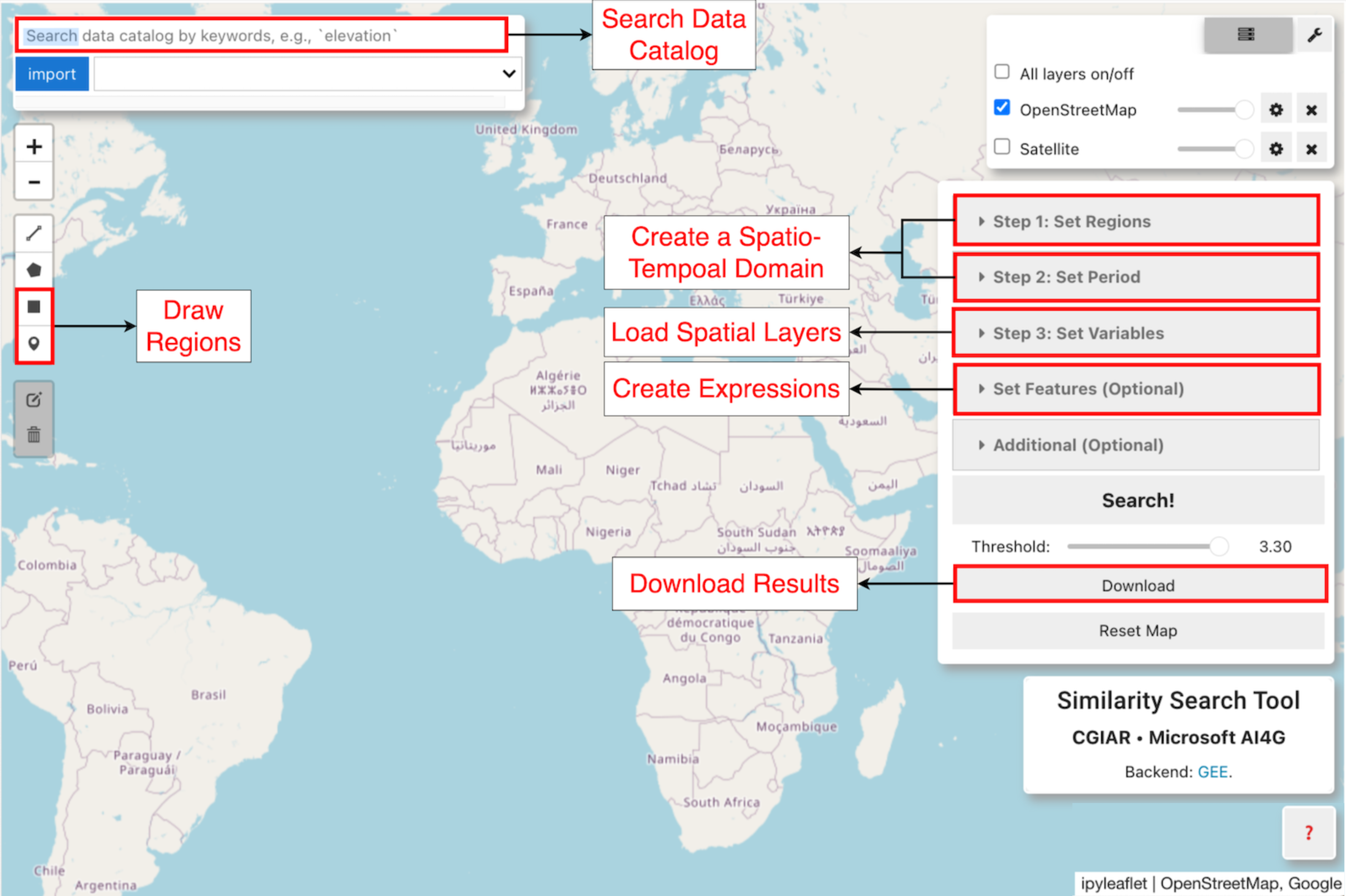}
    \caption{\textbf{Overview of \sims}. The interface includes functionalities such as searching the data catalog, drawing or uploading regions of interest, creating spatio-temporal domains, loading \& visualizing layers from Google Earth Engine, creating custom variable expressions (i.e., features), and downloading the resulting cluster or similarity maps.}
    \label{fig:starter_page}
\end{figure*}

To fill this gap, we developed \sims\footnote{\href{https://github.com/microsoft/sims}{Sims: An Interactive Tool for Geospatial Matching and Clustering}} (Similarity Search), a no-code web tool that enables users to visualize, compare, cluster, and perform similarity searches using spatial data layers over defined regions of interest. \sims is designed to streamline the feature discovery process in tasks such as identifying homogeneous agricultural zones and scaling fertilizer recommendations by finding regions with similar agro-ecological characteristics. By focusing on feature exploration rather than model creation, \sims complements existing spatial modeling tools and helps users quickly identify meaningful spatial variables for geospatial analysis.

Here are our main contributions in this paper:
\begin{itemize}
    \item \sims: We introduce \sims, an open-source web tool for spatial feature exploration.
    \item Case Study: We analyze maize yield patterns in Rwanda using \sims, demonstrating how the tool can help identify distinct yield response zones.
\end{itemize}

\textbf{Audience}. \sims is designed for domain experts and geospatial data scientists and analysts. No coding is required, though familiarity with the GEE data catalog is recommended.

\section{Overview}\label{sec:overview}

\sims is built on top of \textit{geemap}~\cite{geemap}, \textit{ipyleaflet}~\cite{renou2021ipyleaflet}, and \textit{Solara} ~\cite{solara2024github}. It provides two main functionalities: clustering and similarity search. Clustering produces spatially similar sub-regions given a query region, time period, and variables of interest (i.e., a feature profile). In clustering, \sims groups pixels within the query region using \texttt{ee.Clusterer.wekaMeans} where users specify the number of clusters $k$. Unlike static zone maps, \sims provides on-demand, spatially and temporally relevant zones. For similarity search, \sims compares query and reference regions using configurable distance metrics (Euclidean, Manhattan, or Cosine distance), generating a heatmap that highlights areas most similar to the reference region. Both clustering and similarity search produce downloadable raster files for further analysis. \sims can be run locally or deployed with \textit{Solara}~\cite{solara2024github}.

To use the tool, we follow the steps on the right side of Figure \ref{fig:starter_page}. We start by defining a spatio-temporal domain of interest. We can draw geometries directly on the map or upload them as vector files. For similarity search, we need to define two geometries: the ``query'' and ``reference'' regions. Next, we specify the time period by selecting a start and end dates. Once the spatio-temporal domain is set, we can load GEE layers and visualize them on the map as \textit{aliases}. We can also create \textit{features} from the existing aliases using GEE Expressions, for example, the feature ``$ndvi=(nir-red)/(nir+red)$'' can be created from the ``$nir$'' and ``$red$'' aliases. Additionally, we can focus our search on specific Dynamic World~\cite{brown2022dynamic} land cover classes. The feature layers get resampled to a target resolution and stacked into one image before perfoming clustering or similarity search.

\subsection{Clustering}\label{subsec:clustering}

\begin{figure*}[htbp]
    \centering
    \includegraphics[width=0.95\textwidth]{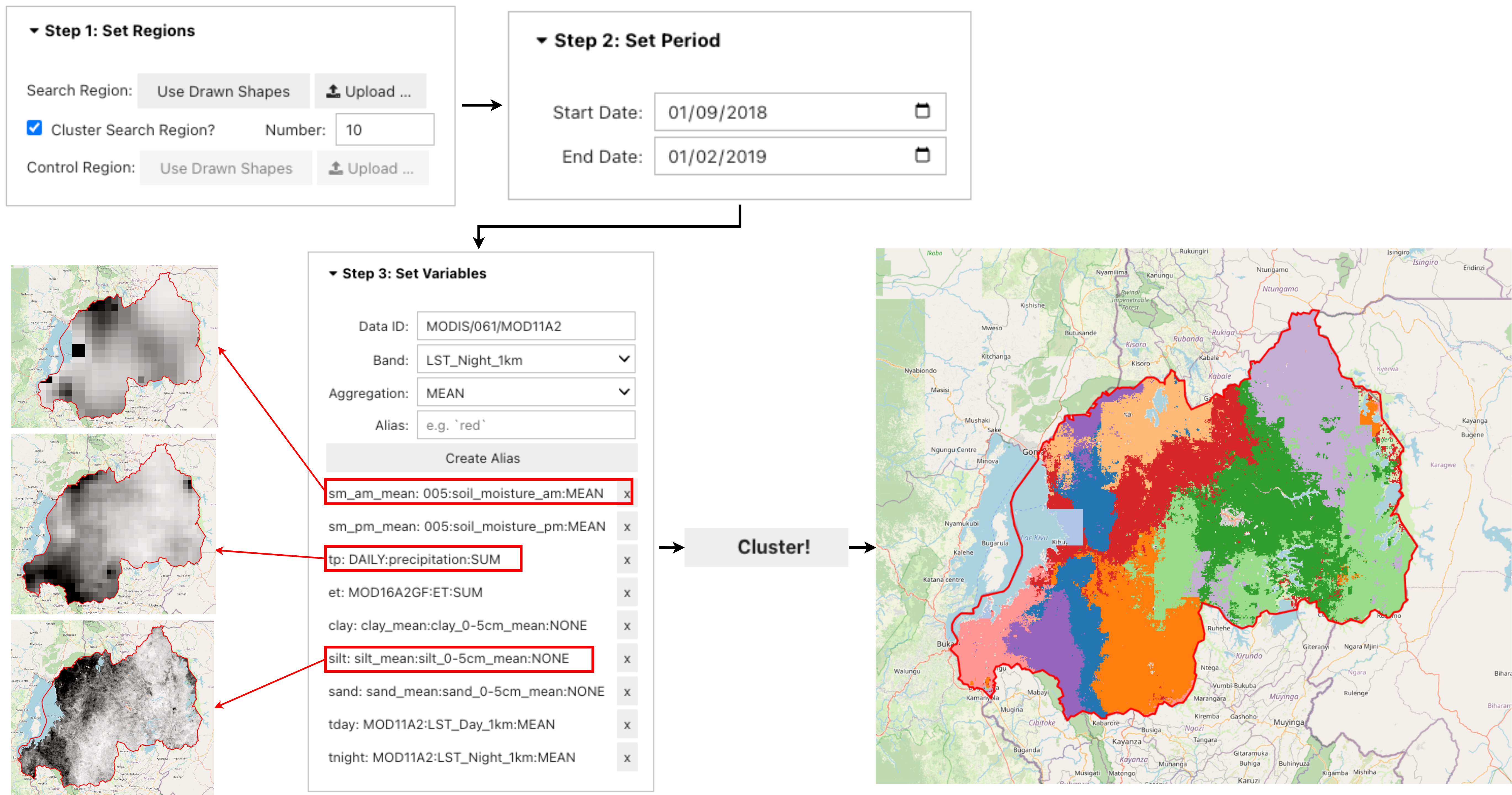}
    \caption{\textbf{Clustering workflow in Sims.} First, define the spatial extent by uploading or drawing a region of interest. Second, set the temporal period for analysis. Third, select and load relevant variables from GEE. Finally, apply clustering to produce distinct zones based on the selected features. The resulting zones represent areas with similar geospatial characteristics.}
    \label{clustering_demo}
\end{figure*}

Figure~\ref{clustering_demo} shows the clustering workflow in \sims. After uploading a boundary file and setting the number of clusters $k$ (i.e., $10$), we set the period of interest (start \& end dates), we select relevant geospatial variables from GEE, including climate (temperature, precipitation), soil moisture and texture, and surface properties. We load the variables as spatial layers with descriptive aliases (e.g., ``tp'', ``et'', ``clay'', etc.) and segment the region into $10$ distinct clusters. The resulting clusters group areas with similar geospatial characteristics, which can be exported as a raster file for further analysis. \sims delivers results within minutes with real-time visualization of the intermediate variables and features.

\subsection{Similarity Search}\label{subsec:similarity_search}

\begin{figure*}[ht]
    \centering
    \includegraphics[width=0.95\textwidth]{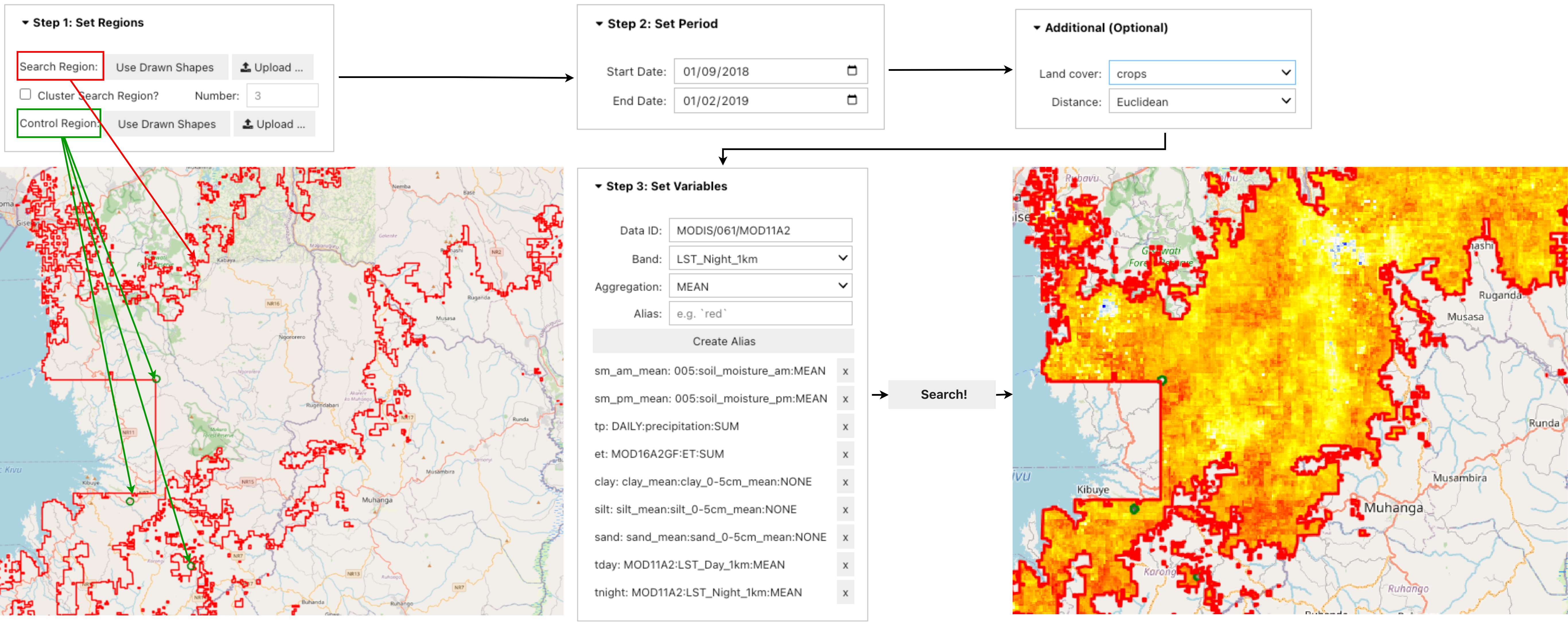}
    \caption{\textbf{Similarity search workflow in Sims.} First, define both search and reference regions by uploading or drawing geometries. Second, set the temporal period and optionally configure land cover masking and distance metrics. Third, select and load relevant variables from GEE. Finally, generate a heat map showing distances between reference and query regions in feature space. The resulting visualization highlights areas that share similar characteristics with the reference region.}
    \label{similarity_search_demo}
\end{figure*}

Figure~\ref{similarity_search_demo} demonstrates the similarity search workflow in \sims. First, we define both the search region (where we want to find similar areas) and reference region (the template area we want to match) by uploading vector files. After setting the temporal period and loading relevant geospatial variables from GEE, we run the similarity search algorithm that reduces the reference region into a vector then calculates the distance to all pixels within the search region. The produced heat map can be further refined using land cover masks and exported as a raster file for downstream analysis.

\section{Case Study: Analyzing Maize Yield Patterns in Rwanda}\label{sec:case_study}
\paragraph*{Background:}
The Excellence in Agronomy (EiA) initiative, of CGIAR, a global research partnership for a food-secure future, addresses global challenges including improving access to appropriate fertilizers for smallholder farmers in Africa, where limited knowledge and resources often lead to sub-optimal crop yields and low profits. EiA has developed a pipeline to generate location-specific agriculture recommendations through its AgWise~\cite{cherneteia} platform. In Rwanda, for example, AgWise provides targeted fertilizer recommendations for key crops such as maize, rice, and potatoes. However, producing timely large-scale recommendations presents significant challenges, particularly in scaling recommendations to regions with very limited field observations. Current methods for testing and identifying optimal fertilizer rates are time consuming and localized, hindering timely deployment of recommendations and widespread impact. 

\paragraph*{Objective:}
To demonstrate how \sims could potentially support such efforts, we analyze simulated maize yield data for Rwanda's January-May cropping seasons between 2005-2015~\cite{maizedatarwanda}, we aim to identify distinct yield response zones. Through \sims' clustering feature, we evaluate different combinations of spatial variables (soil, weather, and agronomic features) to segment regions with significantly different yield responses. Table~\ref{features-table} shows the feature combinations used for each domain, with detailed specifications provided in the appendices (Soil~\ref{appendix:aliases1}, Weather~\ref{appendix:aliases2}, Agronomy~\ref{appendix:aliases3}). For different variable groups and number of clusters ($K$) combinations, we compare the distribution of a medium maturing variety from the simulated maize yield dataset, and assess the significance of the results.

\begin{table}[htbp]
\caption{Features included in each of the different domains considered for the clustering use case, i.e., analyzing maize yield patterns in Rwanda.}
\label{features-table}
\centering
 \resizebox{0.95\linewidth}{!}{%
\begin{tabular}{p{0.5\columnwidth} | p{0.05\columnwidth} | p{0.15\columnwidth} | p{0.15\columnwidth}} 
\toprule
 Feature & Soil & Weather & Agronomy \\ 
 \midrule
 Total Rainfall &  & \checkmark& \checkmark \\ 
 Mean Min \& Max Temperature &  & \checkmark & \checkmark \\
 Mean Relative Humidity &  & \checkmark & \checkmark \\
 Mean Evapotranspiration &  & \checkmark & \checkmark \\
 Soil: Clay \& Sand Content & \checkmark &  & \checkmark \\
 Soil: Nitrogen, Organic Carbon, and pH & \checkmark &  & \checkmark \\
 NDVI &  &  & \checkmark \\ [1ex]
\bottomrule
\end{tabular}
}
\end{table}

\begin{figure*}[htbp]
    \centering
    \includegraphics[width=\textwidth]{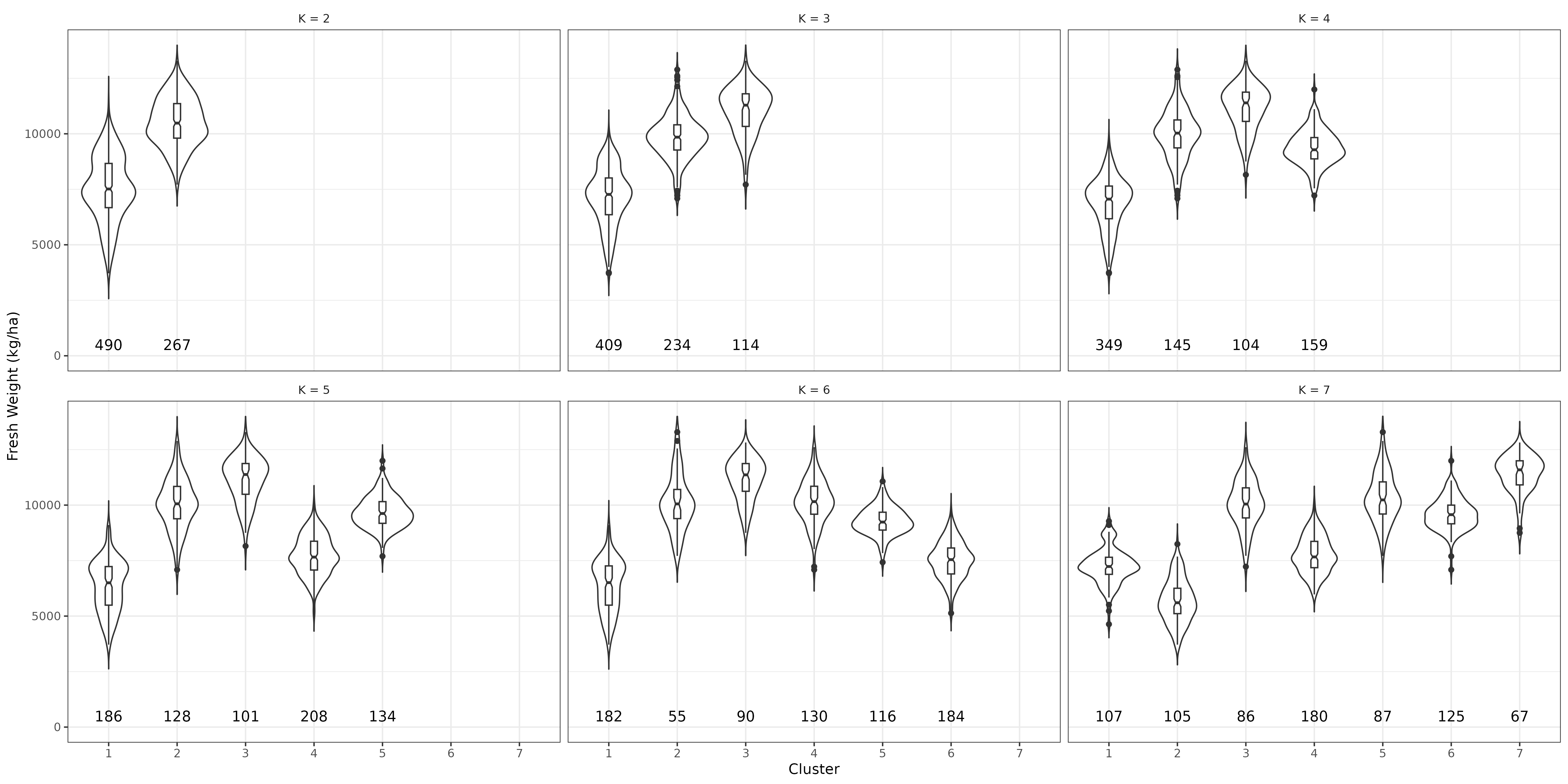}
    \caption{\textbf{Simulated maize yield (kg/ha) distributions for each of the clusters of maize yield patterns in Rwanda.} We experimented with an increasing number of cluster combinations (K) for the set of features from the agronomy domain. The number below each boxplot represent the sample size. Overlapping notches indicate non-significant differences in the yield.}
    \label{boxplots}
\end{figure*}

\paragraph*{Result:} Figure~\ref{boxplots} shows the differences in maize performance across clustered regions and highlights the cluster combinations that best capture yield variations. We found highly significant differences ($p < 2e\text{-}16$) between most cluster pairs at $K=5$, except clusters $2$ and $5$ ($p \approx 1$) and clusters $2$ and $3$ ($p = 1.2e\text{-}09$). These results demonstrate how users can apply the tool to determine optimal variables and cluster numbers for separating primary maize growing regions.

\section{Discussion}\label{sec:discussion}

\subsection{Potential Impact}\label{subsec:potential_impact}

\sims is designed for generic spatial workflows. It supports stacking variables from different data products, creating and inspecting spatial features interactively, and performing operations like clustering and similarity search. Additionally, it requires minimal compute resources; it can be deployed on a small virtual machine instance (i.e., 4GB memory, <10GB storage), as all computational tasks are offloaded to GEE. By allowing users to export results for further analysis, \sims enhances the efficiency of spatial modeling workflows without demanding extensive computational resources.

Beyond specific use cases, \sims is versatile enough for more generic tasks. For example, it can facilitate labeling efforts by allowing users to upload existing labels and identify similar candidates, thereby accelerating the annotation process. In disaster response scenarios, \sims can be used to extend a small known flooded area into a larger flood map by taking advantage of relevant features, such as elevation, to search for similar regions. This capability enables the rapid identification of affected areas and supports timely interventions. We also see its broader potential benefit in accelerating geospatial feature discovery workflows for modeling tasks. By leveraging its direct connection to the GEE backend for data loading and computation, \sims enables faster data sourcing and featurization with a short visual feedback cycle. Making it easier for data scientists to inspect multiple layers and features in the search of strong sources of predictability.

\subsection{Limitations}\label{subsec:limitations}

While \sims offers valuable functionalities for geospatial feature discovery, it has some limitations. The tool requires users to have knowledge of the GEE data catalog to select appropriate layers and features. It currently lacks automated selection capabilities for aliases, optimal number of clusters, and feature expressions. Web sessions are not persistent, causing users to lose their configurations upon exiting. The tool also lacks time-series analysis capabilities, requiring temporal aggregation that may miss important patterns. Additionally, Sims does not warn users about or optimize for large-scale computations (e.g. continent scale), which can lead to inefficiencies and high costs.

\section{Conclusion and Future Work}\label{sec:conclusion}
In spatial modeling, feature discovery is slow because imagery is heavy, data processing is expensive, and visualizing intermediate steps takes time and custom code. These factors create delays, making it difficult to quickly identify useful sources of predictability. To address this, we developed \sims, a no-code tool that enables users to load, transform, and visualize remote sensing layers over regions of interest, create similarity and cluster maps from these layers, and export the results for further analysis, facilitating faster and more accurate spatial modeling.

Looking ahead, there are several ways to address the aforementioned limitations. To enable workflow persistence and sharing, users will be able to export/import \textit{Template} files that captures their configuration (spatio-temporal domain, aliases, and features). To handle large-scale computations efficiently, users can specify a \textit{resolution of interest} when visualizing or downloading data. We plan to implement automated meta-selection capabilities to identify optimal spatial layers and determine ideal cluster numbers, reducing the need for GEE catalog expertise and manual feature selection. We also aim to add alias statistics (e.g., histograms) and support time-series analysis. We aim to engage domain experts and the geospatial community to further build upon \sims for a wide range of applications.

\bibliographystyle{IEEEbib}
\bibliography{strings,main}

\clearpage
\appendix

\section{Soil Aliases \& Features Used in the Clustering Case Study}
\label{appendix:aliases1}

Aliases within \sims are defined in the format:
\begin{quote}
\textit{\{alias\_name\}:\{Product\_ID\}:\{Product\_band\}:\\\{Start\_date\}:\{End\_date\}:\{Aggregation\_function\}}
\end{quote}

\begin{itemize}
    \item \textbf{alias\_name}: A short, descriptive name for the variable, making it easy to reference within the tool. Created by the user.
    \item \textbf{Product\_ID}: The identifier for the specific dataset within GEE.
    \item \textbf{Product\_band}: The specific band within the dataset that contains the data of interest, such as soil moisture or temperature.
    \item \textbf{Start/End\_date}: defines the period of interest.
    \item \textbf{Aggregation\_function}: applied to aggregate an image collection over the defined temporal domain, such as MEAN, SUM, or LAST. Produces a single image.
\end{itemize}

\subsection{Aliases}

\begin{itemize}
    \item \textbf{Soil Clay}
    \begin{itemize}
        \item \begin{tabbing}
            \texttt{\small clay5:soilgrids-isric/clay\_mean:}\\
            \texttt{\small clay\_0-5cm\_mean:}\\
            \texttt{\small 01/01/2010:31/12/2020:LAST}
        \end{tabbing}
        \item \begin{tabbing}
            \texttt{\small clay15:soilgrids-isric/clay\_mean:}\\
            \texttt{\small clay\_5-15cm\_mean:}\\
            \texttt{\small 01/01/2010:31/12/2020:LAST}
        \end{tabbing}
        \item \texttt{\small ...}
        \item \begin{tabbing}
            \texttt{\small clay100:soilgrids-isric/clay\_mean:}\\
            \texttt{\small clay\_60-100cm\_mean:}\\
            \texttt{\small 01/01/2010:31/12/2020:LAST}
        \end{tabbing}
    \end{itemize}

\item \textbf{Soil Sand}
   \begin{itemize}
       \item \begin{tabbing}
           \texttt{\small sand5:soilgrids-isric/sand\_mean:}\\
           \texttt{\small sand\_0-5cm\_mean:}\\
           \texttt{\small 01/01/2010:31/12/2020:LAST}
       \end{tabbing}
       \item \begin{tabbing}
           \texttt{\small sand15:soilgrids-isric/sand\_mean:}\\
           \texttt{\small sand\_5-15cm\_mean:}\\
           \texttt{\small 01/01/2010:31/12/2020:LAST}
       \end{tabbing}
       \item \texttt{\small ...}
       \item \begin{tabbing}
           \texttt{\small sand100:soilgrids-isric/sand\_mean:}\\
           \texttt{\small sand\_60-100cm\_mean:}\\
           \texttt{\small 01/01/2010:31/12/2020:LAST}
       \end{tabbing}
   \end{itemize}
   \item \textbf{Soil Organic Carbon}
   \begin{itemize}
       \item \begin{tabbing}
           \texttt{\small soc5:soilgrids-isric/ocd\_mean:}\\
           \texttt{\small ocd\_0-5cm\_mean:}\\
           \texttt{\small 01/01/2010:31/12/2020:LAST}
       \end{tabbing}
       \item \begin{tabbing}
           \texttt{\small soc15:soilgrids-isric/ocd\_mean:}\\
           \texttt{\small ocd\_5-15cm\_mean:}\\
           \texttt{\small 01/01/2010:31/12/2020:LAST}
       \end{tabbing}
       \item \texttt{\small ...}
       \item \begin{tabbing}
           \texttt{\small soc100:soilgrids-isric/ocd\_mean:}\\
           \texttt{\small ocd\_60-100cm\_mean:}\\
           \texttt{\small 01/01/2010:31/12/2020:LAST}
       \end{tabbing}
   \end{itemize}
   \item \textbf{Soil Total Nitrogen}
   \begin{itemize}
       \item \begin{tabbing}
           \texttt{\small n5:soilgrids-isric/nitrogen\_mean:}\\
           \texttt{\small nitrogen\_0-5cm\_mean:}\\
           \texttt{\small 01/01/2010:31/12/2020:LAST}
       \end{tabbing}
       \item \begin{tabbing}
           \texttt{\small n15:soilgrids-isric/nitrogen\_mean:}\\
           \texttt{\small nitrogen\_5-15cm\_mean:}\\
           \texttt{\small 01/01/2010:31/12/2020:LAST}
       \end{tabbing}
       \item \texttt{\small ...}
       \item \begin{tabbing}
           \texttt{\small n100:soilgrids-isric/nitrogen\_mean:}\\
           \texttt{\small nitrogen\_60-100cm\_mean:}\\
           \texttt{\small 01/01/2010:31/12/2020:LAST}
       \end{tabbing}
   \end{itemize}
   \item \textbf{Soil pH H2O}
   \begin{itemize}
       \item \begin{tabbing}
           \texttt{\small ph5:soilgrids-isric/phh2o\_mean:}\\
           \texttt{\small phh2o\_0-5cm\_mean:}\\
           \texttt{\small 01/01/2010:31/12/2020:LAST}
       \end{tabbing}
       \item \begin{tabbing}
           \texttt{\small ph15:soilgrids-isric/phh2o\_mean:}\\
           \texttt{\small phh2o\_5-15cm\_mean:}\\
           \texttt{\small 01/01/2010:31/12/2020:LAST}
       \end{tabbing}
       \item \texttt{\small ...}
       \item \begin{tabbing}
           \texttt{\small ph100:soilgrids-isric/phh2o\_mean:}\\
           \texttt{\small phh2o\_60-100cm\_mean:}\\
           \texttt{\small 01/01/2010:31/12/2020:LAST}
       \end{tabbing}
   \end{itemize}
\end{itemize}

\subsection{Features}

\begin{itemize}
   \item \texttt{clay:MEAN(clay*)}
   \item \texttt{sand:MEAN(sand*)}
   \item \texttt{soc:MEAN(soc*)}
   \item \texttt{ntot:MEAN(n*)}
   \item \texttt{ph:MEAN(ph*)}
\end{itemize}

\section{Weather Aliases \& Features Used in the Clustering Case Study}
\label{appendix:aliases2}

\subsection{Aliases}

\begin{itemize}
  \item \textbf{Rainfall}
  \begin{itemize}
      \item \begin{tabbing}
          \texttt{\small rain05:UCSB-CHG/CHIRPS/DAILY:}\\
          \texttt{\small precipitation:}\\
          \texttt{\small 01/12/2004:31/07/2005:SUM}
      \end{tabbing}
      \item \begin{tabbing}
          \texttt{\small rain06:UCSB-CHG/CHIRPS/DAILY:}\\
          \texttt{\small precipitation:}\\
          \texttt{\small 01/12/2005:31/07/2006:SUM}
      \end{tabbing}
      \item \texttt{\small ...}
      \item \begin{tabbing}
          \texttt{\small rain15:UCSB-CHG/CHIRPS/DAILY:}\\
          \texttt{\small precipitation:}\\
          \texttt{\small 01/12/2014:31/07/2015:SUM}
      \end{tabbing}
  \end{itemize}
  \item \textbf{Max Temperature}
  \begin{itemize}
      \item \begin{tabbing}
          \texttt{\small tmax05:MODIS/061/MOD11A2:}\\
          \texttt{\small LST\_Day\_1km:}\\
          \texttt{\small 01/12/2004:31/07/2005:MAX}
      \end{tabbing}
      \item \begin{tabbing}
          \texttt{\small tmax06:MODIS/061/MOD11A2:}\\
          \texttt{\small LST\_Day\_1km:}\\
          \texttt{\small 01/12/2005:31/07/2006:MAX}
      \end{tabbing}
      \item \texttt{\small ...}
      \item \begin{tabbing}
          \texttt{\small tmax15:MODIS/061/MOD11A2:}\\
          \texttt{\small LST\_Day\_1km:}\\
          \texttt{\small 01/12/2014:31/07/2015:MAX}
      \end{tabbing}
  \end{itemize}

\item \textbf{Min Temperature}
  \begin{itemize}
      \item \begin{tabbing}
          \texttt{\small tmin05:MODIS/061/MOD11A2:}\\
          \texttt{\small LST\_Night\_1km:}\\
          \texttt{\small 01/12/2004:31/07/2005:MIN}
      \end{tabbing}
      \item \begin{tabbing}
          \texttt{\small tmin06:MODIS/061/MOD11A2:}\\
          \texttt{\small LST\_Night\_1km:}\\
          \texttt{\small 01/12/2005:31/07/2006:MIN}
      \end{tabbing}
      \item \texttt{\small ...}
      \item \begin{tabbing}
          \texttt{\small tmin15:MODIS/061/MOD11A2:}\\
          \texttt{\small LST\_Night\_1km:}\\
          \texttt{\small 01/12/2014:31/07/2015:MIN}
      \end{tabbing}
  \end{itemize}
  \item \textbf{Relative Humidity}
  \begin{itemize}
      \item \begin{tabbing}
          \texttt{\small rhum05:UCSB-CHG/CHIRTS/DAILY:}\\
          \texttt{\small relative\_humidity:}\\
          \texttt{\small 01/12/2004:31/07/2005:MEAN}
      \end{tabbing}
      \item \begin{tabbing}
          \texttt{\small rhum06:UCSB-CHG/CHIRTS/DAILY:}\\
          \texttt{\small relative\_humidity:}\\
          \texttt{\small 01/12/2005:31/07/2006:MEAN}
      \end{tabbing}
      \item \texttt{\small ...}
      \item \begin{tabbing}
          \texttt{\small rhum15:UCSB-CHG/CHIRTS/DAILY:}\\
          \texttt{\small relative\_humidity:}\\
          \texttt{\small 01/12/2014:31/07/2015:MEAN}
      \end{tabbing}
  \end{itemize}
  \item \textbf{Evapotranspiration}
  \begin{itemize}
      \item \begin{tabbing}
          \texttt{\small et10:FAO/WAPOR/2/L1\_AETI\_D:}\\
          \texttt{\small L1\_AETI\_D:}\\
          \texttt{\small 01/12/2009:31/07/2010:MEAN}
      \end{tabbing}
      \item \begin{tabbing}
          \texttt{\small et11:FAO/WAPOR/2/L1\_AETI\_D:}\\
          \texttt{\small L1\_AETI\_D:}\\
          \texttt{\small 01/12/2010:31/07/2011:MEAN}
      \end{tabbing}
      \item \texttt{\small ...}
      \item \begin{tabbing}
          \texttt{\small et15:FAO/WAPOR/2/L1\_AETI\_D:}\\
          \texttt{\small L1\_AETI\_D:}\\
          \texttt{\small 01/12/2014:31/07/2015:MEAN}
      \end{tabbing}
  \end{itemize}
\end{itemize}

\subsection{Features}

\begin{itemize}
       \item \texttt{rain:MEAN(rain*)}
       \item \texttt{tmax:MEAN(tmax*)}
       \item \texttt{tmin:MEAN(tmin*)}
       \item \texttt{rhum:MEAN(rhum*)}
       \item \texttt{et:MEAN(et*)}
\end{itemize}

\section{Agronomy Aliases \& Features Used in the Clustering Case Study}
\label{appendix:aliases3}

\subsection{Aliases}

\begin{itemize}
  \item \textbf{NDVI}
  \begin{itemize}
      \item \begin{tabbing}
          \texttt{\small ndvi05:MODIS/061/MOD13A2:}\\
          \texttt{\small NDVI:}\\
          \texttt{\small 01/12/2004:31/07/2005:MEAN}
      \end{tabbing}
      \item \begin{tabbing}
          \texttt{\small ndvi06:MODIS/061/MOD13A2:}\\
          \texttt{\small NDVI:}\\
          \texttt{\small 01/12/2005:31/07/2006:MEAN}
      \end{tabbing}
      \item \texttt{\small ...}
      \item \begin{tabbing}
          \texttt{\small ndvi15:MODIS/061/MOD13A2:}\\
          \texttt{\small NDVI:}\\
          \texttt{\small 01/12/2014:31/07/2015:MEAN}
      \end{tabbing}
  \end{itemize}
\end{itemize}

\subsection{Features}

\begin{itemize}
       \item \texttt{ndvi:MEAN(ndvi*)}
\end{itemize}

\end{document}